\pdfoutput=1

\documentclass[11pt]{article}

\usepackage[]{acl}

\usepackage{times}
\usepackage{latexsym}

\usepackage[T1]{fontenc}

\usepackage[utf8]{inputenc}

\usepackage{microtype}
\usepackage{multirow}
\usepackage{graphicx} 
\usepackage{float}
\usepackage{times}
\usepackage{latexsym}
\usepackage{CJKutf8}
\usepackage{amsmath}

\usepackage{color}
\usepackage{color}

\usepackage{times}
\usepackage{soul}
\usepackage{url}
\usepackage[utf8]{inputenc}
\usepackage{graphicx}
\usepackage{amssymb,amsmath,amsthm}
\usepackage{booktabs}
\usepackage{algorithm}
\usepackage{algorithmic}
\usepackage{arydshln}
\usepackage{wrapfig}
\usepackage{CJK}
\newtheorem{prop}{Proposition}

\usepackage{mathtools} 

\usepackage{times}
\usepackage{latexsym}

\usepackage[T1]{fontenc}

\usepackage[utf8]{inputenc}

\usepackage{microtype}

\title{MirrorAlign: A Super Lightweight Unsupervised Word Alignment Model\\ via Cross-Lingual Contrastive Learning}

\author{Di Wu \\ Peking University, China \\\texttt{inbath@163.com}
\And
Liang Ding \\ The University of Sydney, Australia \\ \texttt{liangding.liam@gmail.com}
\AND
Shuo Yang \\ iFlytek Research, China \\
\texttt{shuoyang7@iflytek.com}
\And
Mingyang Li \\ Independent Researcher, China\\ \texttt{liamlmy@163.com} \\ 
\texttt{}
}

\begin{document}
\maketitle
\begin{abstract}
Word alignment is essential for the downstream cross-lingual language understanding and generation tasks. Recently, the performance of the neural word alignment models~\cite{garg2019jointly,koehn2019saliency,zenkel-etal-2020-end} has exceeded that of statistical models. However, they heavily rely on sophisticated translation models. In this study, we propose a super lightweight unsupervised word alignment model named \textit{MirrorAlign}, in which a bidirectional symmetric attention trained with a contrastive learning objective is introduced, and an agreement loss is employed to bind the attention maps, such that the alignments follow mirror-like symmetry hypothesis. Experimental results on several public benchmarks demonstrate that our model achieves competitive, if not better, performance compared to the state of the art in word alignment while significantly reducing the training and decoding time on average. Further ablation analysis and case studies show the superiority of our proposed MirrorAlign. Notably, we recognize our model as a pioneer attempt to unify bilingual word embedding and word alignments. Encouragingly, our approach achieves \textit{16.4$\times$ speedup} against GIZA++, and \textit{50$\times$ parameter compression} compared with the Transformer-based alignment methods. We release our code to facilitate the community\footnote{\url{https://github.com/moore3930/MirrorAlign}}.

\end{abstract}

\section{Introduction}

Word alignment, aiming to find the word-level correspondence between a pair of parallel sentences, is a core component of the statistical machine translation ~\cite[SMT]{brown1993mathematics}. It also has benefited several downstream tasks, \emph{e.g.}, computer-aided translation~\cite{dagan1993robust},
semantic role labeling~\cite{kozhevnikov2013cross}, cross-lingual dataset creation~\cite{yarowsky2001inducing}, cross-lingual modeling~\cite{ding-etal-2020-self}, and cross-lingual text generation~\cite{Zan2022BridgingCG}.

Recently, in the era of neural machine translation ~\cite[NMT]{rnnsearch,transformer}, the attention mechanism plays the role of the alignment model in translation system.
Unfortunately, \newcite{koehn2017six} show that attention mechanism may in fact dramatically diverge
with word alignment. 
The works of \newcite{ghader2017does,li2019word} also confirm this finding.

\begin{figure}
    \centering
    \includegraphics[width=0.47\textwidth]{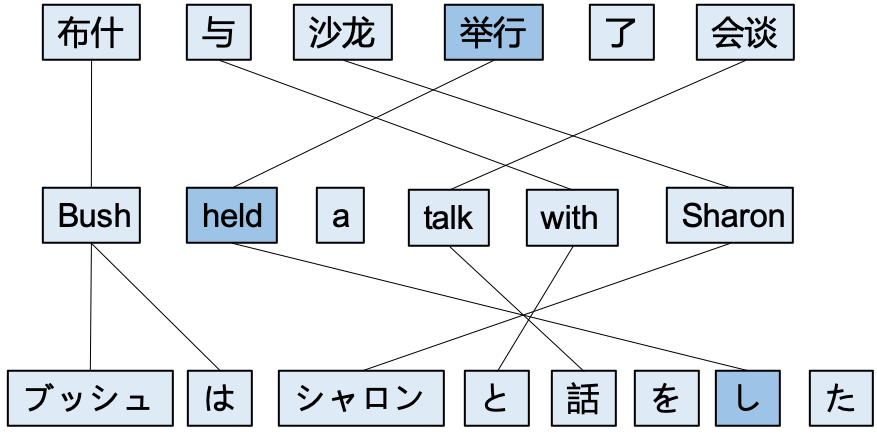}
    \caption{Two examples of word alignment. The upper and bottom cases are the Chinese and Japanese references, respectively.}
    \label{fig:intro}
\end{figure}

Although there are some studies attempt to mitigate this problem, most of them are rely on a sophisticated translation architecture~\cite{garg2019jointly,zenkel-etal-2020-end}. These methods are trained with a translation objective, which computes the probability of each target token conditioned on source tokens and previous target tokens. This will bring tremendous parameters and noisy alignments.
Most recent work avoids the noisy alignment of translation models but employed too much expensive human-annotated alignments~\cite{stengel2019discriminative}. 
Given these disadvantages, simple statistical alignment tools, \emph{e.g.,} FastAlign~\cite{dyer2013simple} and GIZA++~\cite{och2003systematic}\footnote{GIZA++ employs the IBM Model 4 as default setting.}, are still the most representative solutions due to their efficiency and unsupervised fashion. We argue that the word alignment task, intuitively, is much simpler than translation, and thus should be performed before translation rather than inducing alignment matrix with heavy neural machine translation models. For example, the IBM word alignment model, \textit{e.g.}, FastAlign, is the 
prerequisite of SMT. \textit{However, related research about lightweight neural word alignment without NMT is currently very scarce.}


Inspired by cross-lingual word embeddings~\cite[CLWEs]{luong2015bilingual}, we propose to implement a super lightweight unsupervised word alignment model in Section~\ref{sec:method}, named MirrorAlign, which encourages the embedding distance between aligned words to be closer. We also provide the theoretical justification from mutual information perspective for our proposed contrastive learning objective in Section~\ref{sect:training}, demonstrating the reasonableness of our method. Figure~\ref{fig:intro} shows an English sentence, and its corresponding Chinese and Japanese sentences, and their word alignments. The links indicate the correspondence between English$\Leftrightarrow$Chinese and English$\Leftrightarrow$Japanese words.
If the Chinese word \begin{CJK}{UTF8}{gbsn}``举行''\end{CJK} can be aligned to English word ``held'', the reverse mapping should also hold. Specifically, a bidirectional attention mechanism with contrastive estimation is proposed to capture the alignment between parallel sentences. In addition, we employ an agreement loss to constrain the attention maps such that the alignments follow symmetry hypothesis~\cite{liang2006alignment}. 

Our contributions can be summarized as follows:
\begin{itemize}
    \item We propose a super lightweight unsupervised alignment model (MirrorAlign), even merely updating the embedding matrices, achieves better alignment quality on several public benchmark datasets compare to baseline models while preserving comparable training efficiency with FastAlign.
    \item To boost the performance of our model, we design a theoretically and empirically proved bidirectional symmetric attention with contrastive learning objective for word alignment task, in which we introduce extra objective to follow the mirror-like symmetry hypothesis.
    \item Further analysis show that the by-product of our model in training phase has the ability to learn bilingual word representations, which endows the possibility to unify these two tasks in the future.
\end{itemize}

\begin{figure*}
    \centering
    \includegraphics[width=1\textwidth]{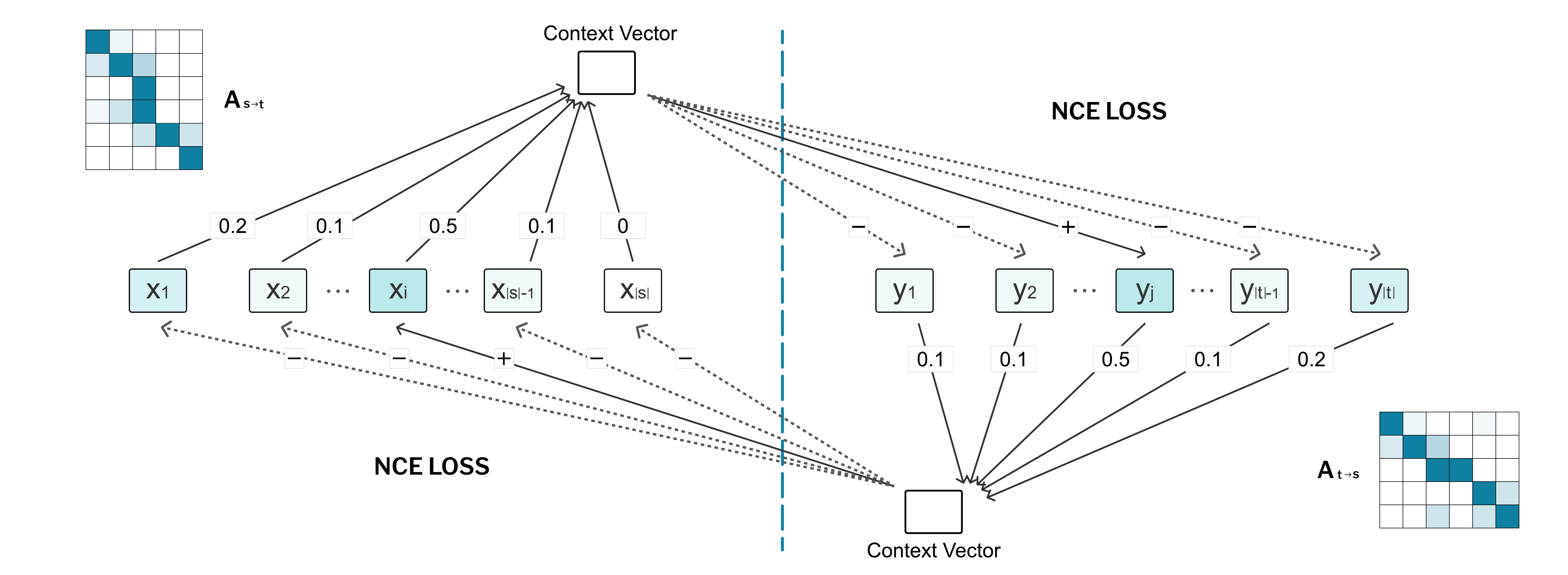}
    \caption{Illustration of MirrorAlign, where a pair of sentences are given as example. Each $x_i$ and $y_j$ are the representation of words in source and target part respectively. Given $y_j$, we can calculate context vector in source part. The NCE training objective is encouraging the dot product of this context vector and $y_j$ to be large. The process in the other direction is consistent. By stacking all of the soft weights, two attention maps $A_{s \rightarrow t}$ and $A_{t \rightarrow s}$ can be produced, which will be bound by an agreement loss to encourage symmetry.
    }
    \label{fig:fig1}
\end{figure*}

\section{Related Work}
Word alignment studies can be divided into two classes:
\paragraph{Statistical Models}
Statistical alignment models directly build on the lexical translation models of \cite{brown1993mathematics}, also known as IBM models. The most popular implementation of this statistical alignment model is FastAlign~\cite{dyer2013simple} and GIZA++ \cite{och2000improved,och2003systematic}. For optimal performance, the training pipeline of GIZA++ relies on multiple iterations of IBM Model 1, Model 3, Model 4 and the HMM alignment model \cite{vogel-etal-1996-hmm}. Initialized with parameters from previous models, each subsequent model adds more assumptions about word alignments. Model 2 introduces non-uniform distortion, and Model 3 introduces fertility. Model 4 and the HMM alignment model introduce relative distortion, where the likelihood of the position of each alignment link is conditioned on the position of the previous alignment link. FastAlign~\cite{dyer2013simple}, which is based on a reparametrization of IBM Model 2, is almost the existing fastest word aligner, while keeping the quality of alignment.

In contrast to GIZA++, our model achieves nearly 15$\times$ speedup during training, while achieving the comparable performance. Encouragingly, our model is at least 1.5$\times$ faster to train than FastAlign and consistently outperforms it.

\paragraph{Neural Models}
Most neural alignment approaches in the literature, such as \citealt{alkhouli2018alignment}, rely on alignments generated by statistical systems that are used as supervision for training the neural systems. These approaches tend to learn to copy the alignment errors from the supervising statistical models. \newcite{zenkel2019adding} use attention to extract alignments from a dedicated alignment layer of a neural model without using any output from a statistical aligner, but fail to match the quality of GIZA++. \newcite{garg2019jointly} represents the current state of the art in word alignment, outperforming GIZA++ by training a single model that is able to both translate and align. This model is supervised with a guided alignment loss, and existing word alignments must be provided to the model during training. \newcite{garg2019jointly} can produce alignments using an end-to-end neural training pipeline guided by attention activations, but this approach underperforms GIZA++. The performance of GIZA++ is only surpassed by training the guided alignment loss using GIZA++ output. \newcite{stengel2019discriminative} introduce a discriminative neural alignment model that uses a dot-product-based distance measure between learned source and target representation to predict if a given source-target pair should be aligned. Alignment decisions are conditioned on the neighboring decisions using convolution. The model is trained using gold alignments. \newcite{zenkel-etal-2020-end} uses guided alignment training, but with large number of modules and parameters, they can surpass the alignment quality of GIZA++. 

They either use translation models for alignment task, which introduces a extremely huge number of parameters (compared to ours), making the training and deployment of the model cumbersome. Or they train the model with the alignment supervision, however, these alignment data is scarce in practice especially for low resource languages. These settings make above approaches less versatile.

Instead, our approach is fully unsupervised at word level, that is, it does not require gold alignments generated by human annotators during training. Moreover, our model achieves comparable performance and is at least 50 times smaller than theirs, i.e., \#Parameters: ~4M (ours) vs. ~200M (above).

\section{Our Approach}
\label{sec:method}
Our model trains in an unsupervised fashion, where the word level alignments are not provided. Therefore, we need to leverage sentence-level supervision of the parallel corpus. To achieve this, we introduce negative sampling strategy with contrastive learning to fully exploit the corpus.
Besides, inspired by the concept of cross-lingual word embedding, we design the model under the following assumption: \textit{If a target token can be aligned to a source token, then the dot product of their embedding vectors should be large.} Figure~\ref{fig:fig1} shows the schema of our approach \textbf{MirrorAlign}.

\subsection{Sentence Representation}
\label{sect:sentence}
For a given source-target sentence pair $(\mathbf{s} ,\mathbf{t})$, $s_i, t_j \in \mathbb{R}^d$ represent the $i$-th and $j$-th word embeddings for the source and target sentences, respectively.  \newcite{luong2015effective,ding-etal-2020-context} illustrate that modelling the neighbour words within the local window helps to understand the current words. Inspired by this, we perform a extremely simple but effective mean pooling operation with the representations of its surrounding words to capture the contextualized information. Padding operation is used to ensure the sequence length. As a result, the final representation of each word can be calculated by element-wisely adding the mean pooling embedding and its original embedding:

\begin{equation}
\begin{split}
x_i = &\textsc{MeanPool}([s_i]^{win}) + s_{i}, \\   
\end{split}
\end{equation}
where $win$ is the pooling window size. We can therefore derive the sentence level representations $( x_1,x_2,...,x_{|s|}), (y_1,y_2,...,y_{|t|})$ for $\mathbf{s}$ and $\mathbf{t}$. In addition to modeling words, modeling structured information (such as syntactic information) may be helpful to enhance the sentence representation~\cite{li-EtAl:2017:Long,marcheggiani-titov:2017:EMNLP2017,ding2019recurrent}, thus improving the word alignment. We leave this exploration for future work.

\subsection{Bidirectional Symmetric Attention}
\label{sect:bi-attention}
Bidirectional symmetric attention is the basic component of our proposed model. The aim of this module is to generate the source-to-target (\textit{aka.} s2t) and target-to-source (\textit{aka.} t2s) soft attention maps.
The details of the attention mechanism: given a source side word representation $x_i$ as query $q_i\in \mathbb{R}^{d}$ and pack all the target tokens together into a matrix $V_t \in \mathbb{R}^{\left | t \right | \times d}$. The attention context can be calculated as:
\begin{equation}
\textsc{Attention}\left(q_i, V_t, V_t\right) = (a_t^{i} \cdot V_t)^\intercal,
\end{equation}
where the vector $a_t^{i} \in \mathbb{R}^{1\times|t|}$ represents the attention probabilities for $q_i$ in source sentence over all the target tokens, in which each element signifies the relevance to the query, and can be derived from:
\begin{equation}
a_t^{i} = \textsc{Softmax} \left(V_t \cdot q_i \right)^\intercal.
\end{equation}
For simplicity, we denote the attention context of $q_i$ in the target side as $att_t(q_i)$.  
s2t attention map $A_{s,t} \in \mathbb{R}^{|s| \times |t|}$ is constructed by stacking the probability vectors $a_t^{i}$ corresponding to all the source tokens.

Reversely, we can obtain t2s attention map $A_{t,s}$ in a symmetric way. Then, these two attention matrices $A_{s,t}$ and $A_{t,s}$ will be used to decode alignment links. Take s2t for example, given a target token, the source token with the highest attention weight is viewed as the aligned word.

\subsection{Agreement Mechanism}
\label{sect:agreement}
Intuitively, the two attention matrices $A_{s,t}$ and $A_{t,s}^{T}$ should be very close. However, the attention mechanism suffers from symmetry error in different direction~\cite{koehn2017six}. 

To bridge this discrepancy, we introduce agreement mechanism~\cite{liang2006alignment}, acting like a mirror that precisely reflects the matching degree between $A_{s,t}$ and $A_{t,s}$, which is also empirically confirmed in machine translation~\cite{levinboim-etal-2015-model}. In particular, we use an agreement loss to bind above two matrices:
\begin{equation}
\mathcal{L}\textit{oss}_{disagree}=\sum_i\sum_j(A^{s,t}_{i,j} - A^{t,s}_{j,i})^2.
\end{equation}

In Section~\ref{subsec:ablation}, we empirically show this agreement can be complementary to the bidirectional symmetric constraint, demonstrating the effectiveness of this component.

\begin{table*}
\centering
\scalebox{1}{
\begin{tabular}{l|ccc|ccc|ccc}
\hline\hline
\textbf{Method} & \textbf{EN-FR} & \textbf{FR-EN} & \textbf{sym} & \textbf{RO-EN} & \textbf{EN-RO} & \textbf{sym} & \textbf{DE-EN} & \textbf{EN-DE} & \textbf{sym}\\ 
\hline
NNSA & 22.2 & 24.2 & 15.7 & 47.0 & 45.5 & 40.3 & 36.9 & 36.3 & 29.5 \\
FastAlign & 16.4 & 15.9 & 10.5 & 33.8 & 35.5 & 32.1 & 28.4 & 32.0 & 27.0 \\
MirrorAlign & \textbf{15.3} & \textbf{15.6} & \textbf{9.2} & 34.3 & \textbf{35.2} & \textbf{31.6} & 31.1 & \textbf{28.0} & \textbf{24.8} \\
\hline
\hline
\end{tabular}
}
\caption{\label{table-2} AER of each method in different direction. ``sym'' means grow-diag symmetrization.}
\end{table*}

\subsection{Training Objective and Theoretical Justification}
\label{sect:training}

Suppose that $(q_i, att_t(q_i))$ is a pair of s2t word representation and corresponding attention context sampled from the joint distribution $p_t(q, att_t(q))$ (hereinafter we call it a positive pair), the primary objective of the s2t training is to maximize the alignment degree between the elements within a positive pair. Thus, we first define an alignment function by using the $\mathrm{sigmoid}$ inner product as:
\begin{equation}
    \textsc{Align}(q, att_t(q)) = \sigma(\langle q, att_t(q)\rangle),
\end{equation}
where $\sigma(\cdot)$ denotes the $\mathrm{sigmoid}$ function and $\langle\cdot,\cdot\rangle$ is the inner product operation. However, merely optimizing the alignment of positive pairs ignores important
positive-negative relation knowledge~\cite{mikolov2013distributed}. 

To make the training process more informative, we reform the overall objective in the contrastive learning manner~\cite{oord2018representation,saunshi2019theoretical} with Noise Contrastive Estimation (NCE) loss~\cite{mikolov2013distributed}, which has been widely used in many NLP tasks~\cite{Xiong2021ApproximateNN,Gao2021SimCSESC,Wang2022ACC}. Specifically, we first sample $k$ negative word representations $q_j$\footnote{In the contrastive learning setting, $q_j$ and $att_t(q_i)$ can be sampled from different sentences. If $q_j$ and $att_t(q_i)$ are from the same sentence, $i\neq j$; otherwise, $j$ can be a random index within the sentence length. For simplicity, in this paper, we use $q_j$ where $i\neq j$ to denote the negative samples, although with a little bit ambiguity.}  from the margin $p_t(q)$. Then, we can formulate the overall NCE objective as following:

\begin{equation}\label{eq:obj}
\begin{aligned}
    &\mathcal{L}\textit{oss}_{s \rightarrow t}^i = -\mathop{\mathbb{E}}_{\{att_t(q_i), q_i, q_j\}}~[ \log \\
    &\frac{\textsc{Align}(q_i, att_t(q_i))}{\textsc{Align}(q_i, att_t(q_i))+\sum_{j=1}^k \textsc{Align}(q_j, att_t(q_i))}],
\end{aligned}
\end{equation}

It is evident that the objective in Eq. (\ref{eq:obj}) explicitly encourages the alignment of positive pair $(q_i, att_t(q_i))$ while simultaneously separates the negative pairs $(q_j, att_t(q_i))$. 

Moreover, a direct consequence of minimizing Eq. (\ref{eq:obj}) is that the optimal estimation of the alignment between the representation and attention context is proportional to the ratio of joint distribution and the product of margins $\frac{p_t(q, att_t(q))}{p_t(q)\cdot p_t(att_t(q))}$ which is the point-wise mutual information, and we can further have the following proposition with repect to the mutual information:
\begin{prop}\label{pro:milb}
    The mutual information between the word representation $q$ and its corresponding attention context $att_t(q)$ is lower-bounded by the negative $\mathcal{L}\textit{oss}_{s \rightarrow t}^i$ in Eq. (\ref{eq:obj}) as:
    \begin{equation}\label{eq:milb}
    I(q, att_t(q)) \geq \log(k) - \mathcal{L}\textit{oss}_{s \rightarrow t}^i,
    \end{equation}
    where $k$ is the number of the negative samples.
\end{prop}

The detailed proof can be found in \cite{oord2018representation}. Proposition \ref{pro:milb} indicates that the lower bound of the mutual information $I(q, att_t(q))$ can be maximized by achieving the optimal NCE loss, which provides theoretical guarantee for our proposed method.

Our training schema over parallel sentences is mainly inspired by the bilingual skip-gram model~\cite{luong2015bilingual} and invertibility modeling~\cite{levinboim-etal-2015-model}. 
Therefore, the ultimate training objective should consider both forward ($s \rightarrow t$) and backward ($t \rightarrow s$) direction, combined with the mirror agreement loss. Technically, the final training objective is:
\begin{equation}
\begin{aligned}
    \mathcal{L}\textit{oss} &= \sum_{i}^{\left| t \right|} \mathcal{L}\textit{oss}_{s \rightarrow t}^{i} + \sum_{j}^{\left| s \right|} \mathcal{L}\textit{oss}_{t \rightarrow s}^{j}\\ &+ \alpha \cdot \mathcal{L}\textit{oss}_{disagree},
\end{aligned}
\end{equation}
where $\mathcal{L}\textit{oss}_{s \rightarrow t}$ and $\mathcal{L}\textit{oss}_{t \rightarrow s}$ are symmetrical and $\alpha$ is a loss weight to balance the likelihood and disagreement loss.


\begin{table}
\centering
\scalebox{0.88}{
\begin{tabular}{lccc}
\hline\hline
\textbf{Model} & \textbf{EN-FR} & \textbf{RO-EN} & \textbf{DE-EN} \\ 
\hline
Naive Attention & 31.4 & 39.8 & 50.9 \\
NNSA & 15.7 & 40.3 & - \\
FastAlign & 10.5 & 32.1 & 27.0 \\
\textbf{MirrorAlign} & \textbf{9.2} & \textbf{31.6} & \textbf{24.8} \\
\hline
\cite{zenkel-etal-2020-end} & 8.4 & 24.1 & 17.9 \\
\cite{garg2019jointly} & 7.7 & 26.0 & 20.2 \\
GIZA++ & 5.5 & 26.5 & 18.7 \\
\hline
\hline
\end{tabular}}
\caption{\label{table-1}Alignment performance (with grow-diagonal heuristic) of each model.
}
\end{table}

\begin{figure*}
    \centering
    \includegraphics[width=0.9\textwidth]{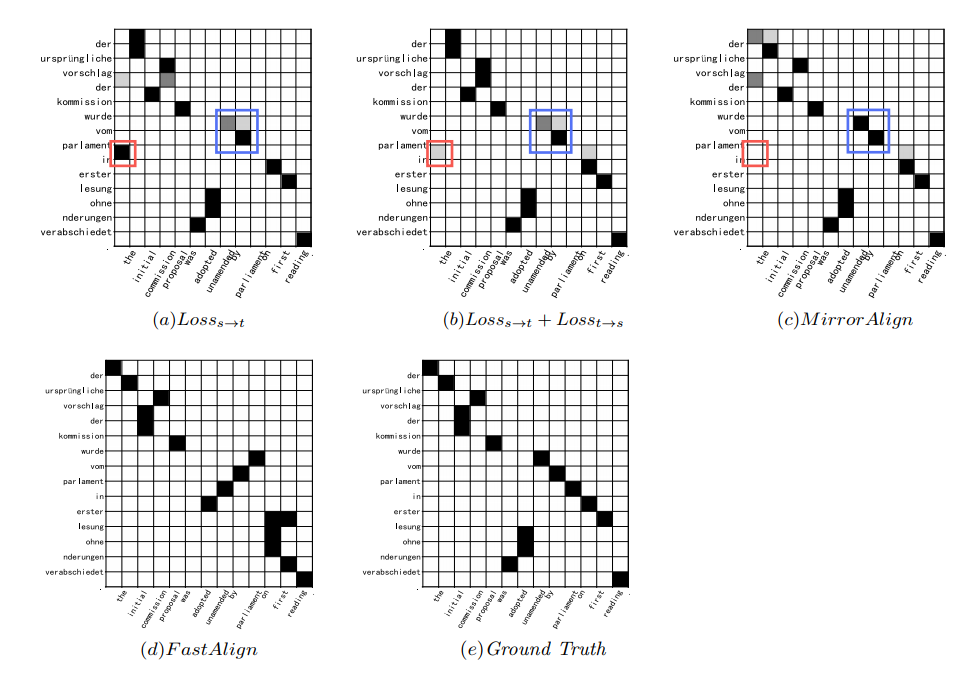}
    \caption{An visualized alignment example. (a-c) illustrate the effects when gradually adding the symmetric component, (d) shows the result of FastAlign, and (e) is the ground truth. The more emphasis is placed on the symmetry of the model, the better the alignment results model achieved. Meanwhile, as depicted, the results of the attention map become more and more diagonally concentrated.}
    \label{fig:appendix-2}
\end{figure*}

\section{Experiments}

\subsection{Datasets and Evaluation Metrics}
We perform our method on three widely used datasets: English-French (\textbf{EN-FR}), Romanian-English (\textbf{RO-EN}) and German-English (\textbf{DE-EN}). Training and test data for \textbf{EN-FR} and \textbf{RO-EN} are from NAACL 2003 share tasks~\cite{mihalcea2003evaluation}. For \textbf{RO-EN}, we add Europarl v8 corpus, increasing the amount of training data from 49K to 0.4M. For \textbf{DE-EN}, we use the Europarl v7 corpus as training data and test on the gold alignments.
All above data are lowercased and tokenized by Moses. The evaluation metrics are Precision, Recall, F-score (F1) and Alignment Error Rate (AER).

\subsection{Baseline Methods}
Besides two strong statistical alignment models, i.e. FastAlign and GIZA++, we also compare our approach with neural alignment models where they induce alignments either from the attention weights or through feature importance measures.
\label{sec:baseline-Methods}
\paragraph{FastAlign} One of the most popular statistical method which log-linearly reparameterize the IBM model 2 proposed by~\cite{dyer2013simple}.

\paragraph{GIZA++} A statistical generative model~\cite{och2003systematic}, in which parameters are estimated using the Expectation-Maximization (EM) algorithm, allowing it to automatically extract bilingual lexicon from parallel corpus.

\paragraph{NNSA} A unsupervised neural alignment model proposed by~\cite{legrand2016neural}, which applies an aggregation operation borrowed from the computer vision to design sentence-level matching loss. In addition to the raw word indices, following three extra features are introduced: distance to the diagonal, part-of-speech and unigram character position. To make a fair comparison, we report the result of raw feature in NNSA.


\paragraph{Naive Attention} Averaging all attention matrices in the Transformer architecture, and selecting the source unit with the maximal attention value for each target unit as alignments. We borrow the results reported in ~\cite{zenkel2019adding} to highlight the weakness of such naive version, where significant improvement are achieved after introducing an extra alignment layer.

\paragraph{Others}  \newcite{garg2019jointly} and \newcite{zenkel-etal-2020-end} represent the current developments in word alignment, which both outperform GIZA++. However, They both implement the alignment model based on a sophisticated translation model. Further more, the former uses the output of GIZA++ as supervision, and the latter introduces a pre-trained state-of-the-art neural translation model. It is unfair to compare our results directly with them. We report them in Table~\ref{table-1} as references.

\subsection{Setup}
For our method (MirrorAlign), all the source and target embeddings are initialized by Xavier method~\cite{glorot2010understanding}. The embedding size $d$ and pooling window size are set to 256 and 3, respectively. The hyper-parameters $\alpha$ 
is tested by grid search from 0.0 to 1.0 at 0.1 intervals. 
For FastAlign, we train it from scratch by the open-source pipeline\footnote{https://github.com/lilt/alignment-scripts}. Also, we report the results of NNSA and machine translation based model (Section~\ref{sec:baseline-Methods}).
All experiments of MirrorAlign are run on 1 \verb|Nvidia P40| GPU. The CPU model is Intel(R) Xeon(R) CPU E5-2620 v3 @ 2.40GHz. Both FastAlign and MirrorAlign take nearly half a hour to train one million samples.



\subsection{Main Results}
Table~\ref{table-1} summarizes the AER of our method over several language pairs. Our model outperforms all other baseline models. Comparing to FastAlign, we achieve 1.3, 0.5 and 2.2 AER improvements on \textbf{EN-FR}, \textbf{RO-EN}, \textbf{DE-EN} respectively.


Notably, our model exceeds the naive attention model in a big margin in terms of AER (ranging from 8.2 to 26.1) over all language pairs. We attribute the poor performance of the straightforward attention model (translation model) to its contextualized word representation. For instance, when translating a verb, contextual information will be paid attention to determine the form (\emph{e.g.,} tense) of the word, that may interfere the word alignment.

Experiment results in different alignment directions can be found in Table~\ref{table-2}. The grow-diag symmetrization benifits all the models.

\subsection{Speed Comparison}
Take the experiment on EN-FR dataset as an example, MirrorAlign converges to the best performance after running 3 epochs and taking 14 minutes totally, where FastAlign and GIZA++ cost 21 and 230 minutes, respectively, to achieve the best results. Notably, the time consumption will rise dozens of times in neural translation fashion.

\subsection{Ablation Study}
\label{subsec:ablation}
To further explore the effects of several components (\emph{i.e.,} bidirectional symmetric attention, agreement loss) in our MirrorAlign, we conduct an ablation study. Table~\ref{table-3} shows the results on \textbf{EN-FR} dataset. When the model is trained using only $\mathcal{L}\textit{oss}_{s\rightarrow t}$ or $\mathcal{L}\textit{oss}_{t \rightarrow s}$ as loss functions, the AER of them are quite high (20.9 and 23.3). As expected, combined loss function improves the alignment quality significantly (14.1 AER). It is noteworthy that with the rectification of agreement mechanism, the final combination achieves the best result (9.2 AER), indicating that the agreement mechanism is the most important component in MirrorAlign.

To better present the improvements brought by adding each component, we visualize the alignment case in Figure~\ref{fig:appendix-2}. As we can see, each component is complementary to others, that is, the attention map becomes more diagonally concentrated after adding the bidirectional symmetric attention and the agreement constraint.

\begin{table}
\centering
\scalebox{1}{
\begin{tabular}{lcccc}
\hline\hline

\textbf{Setup} & \textbf{P} & \textbf{R} & \textbf{F1} & \textbf{AER} \\ 
\hline
$\mathcal{L}\textit{oss}_{s \rightarrow t}$ & 74.9 & 86.0 & 80.4 & 20.9 \\
$\mathcal{L}\textit{oss}_{t \rightarrow s}$ & 71.9 & 85.3 & 77.3 & 23.3 \\
$\mathcal{L}\textit{oss}_{s \leftrightarrow t}$ & 81.5 & \textbf{90.1} & 86.1 & 14.1 \\
MirrorAlign & \textbf{91.8} & 89.1 & \textbf{90.8} & \textbf{9.2} \\
\hline
\hline
\end{tabular}}
\caption{\label{table-3} Ablation results on EN-FR dataset.}
\end{table}

\section{Analysis}
\begin{figure}
     \centering
     \includegraphics[width=0.47\textwidth]{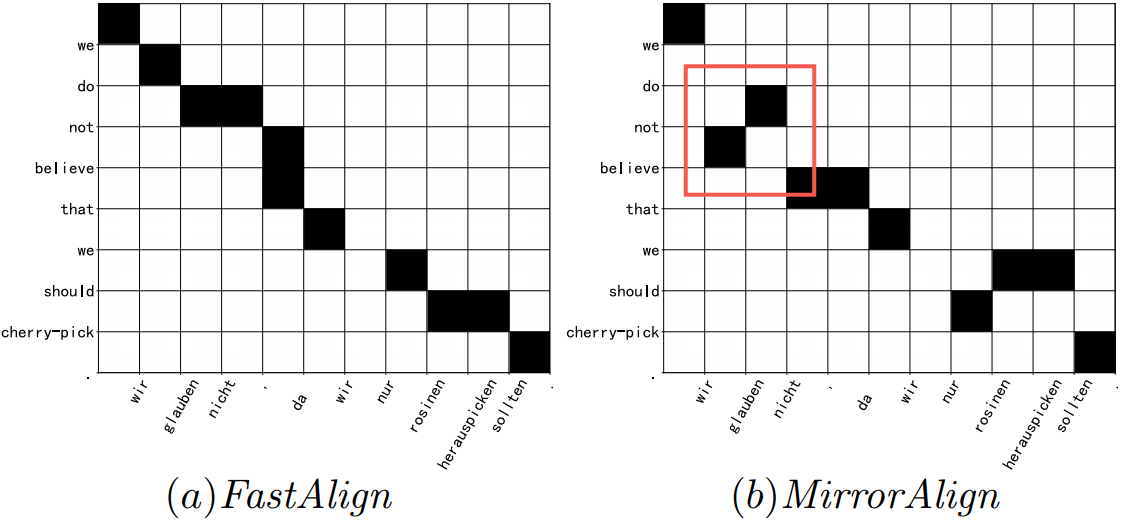}
        \caption{Example of the DE-EN alignment. (a) is the result of FastAlign, and (b) shows result of our model, which is closer to the gold alignment. The horizontal axis shows German sentence ``\textit{wir glauben nicht , da wir nur rosinen herauspicken sollten .}'', and the vertical axis shows English sentence ``\textit{we do not believe that we should cherry-pick .}''.}
        \label{fig:alignments}
\end{figure}

\paragraph{Alignment Case Study}
Figure~\ref{fig:alignments} shows an alignment example. Our model correctly aligns ``\emph{do not believe}'' in English to ``\emph{glauben nicht}'' in German. 
Our model, based on word representation, makes better use of semantics to accomplish alignment such that inverted phrase like ``\emph{glauben nicht}'' can be well handled. Instead, FastAlign, relied on the positional assumption\footnote{A feature $h$ of position is introduced in FastAlign to encourage alignments to occur around the diagonal. $h(i,j,m,n)=-\left | \frac{i}{m} - \frac{j}{n} \right |$, $i$ and $j$ are source and target indices and $m$ and $n$ are the length of sentences pair.}, fails here.


\paragraph{Word Embedding Clustering}
To further investigate the effectiveness of our model, we also analyze the word embeddings learned by our model. In particular, following~\cite{collobert2011natural}, we show some words together with its nearest neighbors using the Euclidean distance between their embeddings.
Table~\ref{table-4} shows some examples to demonstrates that our learned representations possess a clearly clustering structure bilingually and monolingually. 
We attribute the better alignment results to the ability of our model that could learn bilingual word representation.

\begin{table}
\centering
\scalebox{0.8}{
\begin{tabular}{cc|cc}
\hline\hline
\multicolumn{2}{c|}{china} & \multicolumn{2}{c}{distinctive}\\
\hline
\textbf{EN} & \textbf{DE} & \textbf{EN} & \textbf{DE}\\
\hline
china & chinas & distinctive & unverwechselbaren \\
chinese & china & distinct & besonderheiten\\
china's & chinesische & peculiar & markante \\
republic & chinesischer & differences & charakteristische \\
china' & chinesischem & diverse & einzelnen \\
\hline\hline
\multicolumn{2}{c|}{cat} & \multicolumn{2}{c}{love} \\
\hline
\textbf{EN} & \textbf{DE} & \textbf{EN} & \textbf{DE} \\
\hline
cat & hundefelle & love & liebe \\
dog & katzenfell & affection & liebt \\
toys & hundefellen & loved & liebe \\
cats & kuchen & loves & lieben \\
dogs & schlafen & passion & lieb \\
\hline\hline
\end{tabular}}
\caption{\label{table-4} Top 5 nearest English (\textbf{EN}) and German (\textbf{DE}) words for each of the following words: \textit{china}, \textit{distinctive}, \textit{cat}, and \textit{love}.} 
\end{table}

\section{Conclusion and Future Work}
\label{ssec:conclusion}

In this paper, we presented a super lightweight neural alignment model, named MirrorAlign, that has achieved better alignment performance compared to FastAlign and other existing neural alignment models while preserving training efficiency. We empirically and theoretically show its effectiveness over several language pairs. In the future, we would further explore the relationship between CLWEs and word alignments. A promising attempt is using our model as a bridge to unify cross-lingual embeddings and word alignment tasks.


\bibliography{anthology,acl_latex}
\bibliographystyle{acl_natbib}




\end{document}